\documentclass[pdflatex,sn-mathphys-num]{sn-jnl}% Math and Physical Sciences Numbered Reference Style 
\usepackage{graphicx}%
\usepackage{multirow}%
\usepackage{subcaption}
\usepackage{comment}
\usepackage{amsmath,amssymb,amsfonts}%
\usepackage{amsthm}%
\usepackage{mathrsfs}%
\usepackage{tabularx}
\usepackage[title]{appendix}%
\usepackage{textcomp}%
\usepackage{manyfoot}%
\usepackage{booktabs}%
\usepackage{algorithm}%
\usepackage{algorithmicx}%
\usepackage{algpseudocode}%

\usepackage{listings}%
\usepackage{booktabs} % For prettier tables
\usepackage[table]{xcolor} % For row colors
\usepackage{siunitx} % For aligning at decimal points
\usepackage{threeparttable}%

\theoremstyle{thmstyleone}%
\theoremstyle{thmstyletwo}%

\theoremstyle{thmstylethree}%

\begin{document}

\title[Article Title]{Indoor scene recognition from images under visual corruptions}


\author*[1]{\fnm{Willams} \sur{de Lima Costa}}\email{wlc2@cin.ufpe.br}

\author[2]{\fnm{Raul} \sur{Ismayilov}}

\author[2]{\fnm{Nicola} \sur{Strisciuglio}}

\author[2]{\fnm{Estefania} \sur{Talavera Martinez}}

\affil*[1]{\orgdiv{Voxar Labs, Centro de Informática}, \orgname{Universidade Federal de Pernambuco}, \orgaddress{\street{Av. Jornalista Aníbal Fernandes, s/n}, \city{Recife}, \postcode{50740-560},
\country{Brazil}}}

\affil[2]{\orgdiv{Data Management and Biometrics Group}, \orgname{University of Twente}, \orgaddress{\street{Drienerlolaan 5}, \city{Enschede}, \postcode{7522 NB},
\country{The Netherlands}}}


\abstract{The classification of indoor scenes is a critical component in various applications, such as intelligent robotics for assistive living. While deep learning has significantly advanced this field, models often suffer from reduced performance due to image corruption. This paper presents an innovative approach to indoor scene recognition that leverages multimodal data fusion, integrating caption-based semantic features with visual data to enhance both accuracy and robustness against corruption. We examine two multimodal networks that synergize visual features from CNN models with semantic captions via a Graph Convolutional Network (GCN). Our study shows that this fusion markedly improves model performance, with notable gains in Top-1 accuracy when evaluated against a corrupted subset of the Places365 dataset. Moreover, while standalone visual models displayed high accuracy on uncorrupted images, their performance deteriorated significantly with increased corruption severity. Conversely, the multimodal models demonstrated improved accuracy in clean conditions and substantial robustness to a range of image corruptions. These results highlight the efficacy of incorporating high-level contextual information through captions, suggesting a promising direction for enhancing the resilience of classification systems.}

\maketitle
\subsection*{Availability of data and material}
The dataset generated during this research will be made publicly available at\\ \url{https://places-corrupted.github.io} upon the publication of this manuscript.
\subsection*{Competing interests}
The authors have no competing interests to declare.
\subsection*{Funding}
The authors have no funding to report.
\subsection*{Authors' contributions}
\textbf{Willams de Lima Costa}: Conceptualization, Methodology, Validation, Writing - Original Draft, Supervision, Project administration. \textbf{Raul Ismayilov}: Conceptualization, Methodology, Software, Validation, Investigation, Data Curation. \textbf{Nicola Strisciuglio}: Writing - Review \& Editing, Project administration. \textbf{Estefania Talavera Martinez}: Conceptualization, Methodology, Writing - Review \& Editing, Supervision.

\newpage
\section{Introduction}\label{sec:introduction}

\begin{figure}[h]
\centering
\begin{subfigure}{0.5\columnwidth}
\centering
\includegraphics[width=0.30\columnwidth]{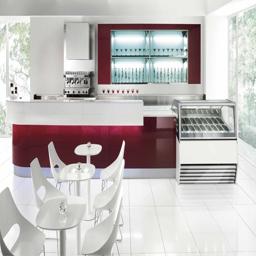}
\hfill
\includegraphics[width=0.30\columnwidth]{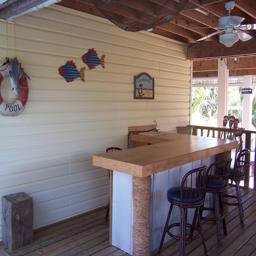}
\hfill
\includegraphics[width=0.30\columnwidth]{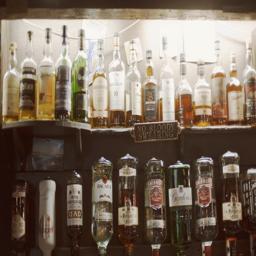}
\caption{bar}
\end{subfigure}
\begin{subfigure}{0.5\columnwidth}
\centering
\includegraphics[width=0.30\columnwidth]{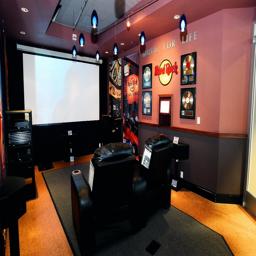}
\hfill
\includegraphics[width=0.30\columnwidth]{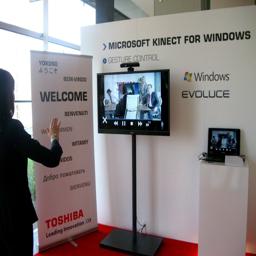}
\hfill
\includegraphics[width=0.30\columnwidth]{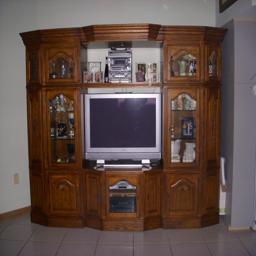}
\caption{home theater}
\end{subfigure}
\begin{subfigure}{0.5\columnwidth}
\centering
\includegraphics[width=0.30\columnwidth]{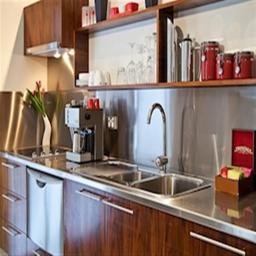}
\hfill
\includegraphics[width=0.30\columnwidth]{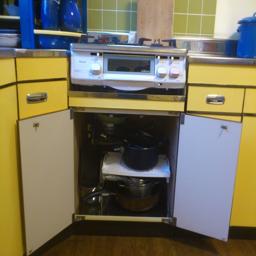}
\hfill
\includegraphics[width=0.30\columnwidth]{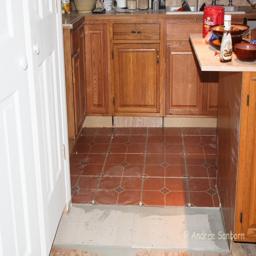}
\caption{kitchen}
\end{subfigure}
\caption{Examples of indoor scenes with high variability within Places365 \citep{zhou2017places}. In these samples, we can note that these images share little to no features with each other.}
\label{fig:variability}
\end{figure}

Indoor scene recognition consists of recognizing indoor locations from features extracted from the environment. This task can be complex due to the high variability and ambiguity of indoor scenes, as we exemplify in \autoref{fig:variability}, where we can observe sample images for a set of given scenes with great variability of appearances or arrangements. Several approaches have risen in the past to solve this problem and are able to achieve significant performance for this task through the incorporation of information that goes beyond image-based features, such as depth maps or 3D reconstructions of the scene \citep{10.1007/978-3-642-24088-1_14, 10.1016/j.isatra.2020.10.023}.

Although many of these state-of-the-art techniques work well on selected datasets and controlled scenarios, visual variations might happen in real environments when capturing images. Common corruptions in the gathered images harms the operational range of such techniques, since existing approaches depends on visual appearance (despite multiple modalities) and expects high-quality input images. These images might contain artifacts that are added during the process of capturing images, such as, for example, motion-blurred images due to the shakiness of hands and arms when holding cameras or smartphones; soft or strong noise due to poor lighting or camera settings during shuttle time; compression methods such as JPEG compressions when sharing images to social networks or other social media, reducing overall image quality and adding artifacts. 

Therefore, a new question arises, as we are not only interested in describing an indoor scene but also in doing so especially in operational settings where images can present any type of unexpected variations or visual corruptions. In response to this challenge, we propose a shift in mindset from existing approaches by seeking a methodology that can generate robust descriptions of images that are sufficient for both clean and corrupted scenarios.

In this work, we propose a two-stream encoding process for indoor scene recognition. This multimodal approach helps improve the robustness of this task by combining different descriptions from the scene. We call the first stream \textit{high-level image description}, in which we generate text descriptions of the scene as if a human was describing what they are seeing. The second stream is the \textit{low-level image description}, in which we extract visual features from the scene using encoder networks (e.g.,  ResNet-50). By fusing these two descriptions in a late fusion approach, we show that combining text and vision descriptions improved classification accuracy, even for highly corrupted images. 

The contributions of this work are three-fold:
\begin{itemize}
    \item We show that combining text and vision is a promising way of improving robustness with respect to common visual corruptions. Although the focus of this work is indoor scene recognition, the same principle could be expanded to other tasks. 
    \item We introduce and make publicly available the Places148-corrupted benchmark, a novel dataset for indoor scene recognition on corrupted samples.
    \item We also present baseline results on Places148-corrupted, which we expect will contribute to further research in the field of indoor scene recognition.
\end{itemize}

The paper is organized as follows: In Section 2, we discuss related work on indoor scene recognition and explore the limitations of current state-of-the-art methods. Section 3 presents the methodology proposed in this study. In Section 4, we detail the experimentation process used to evaluate our method, including a description of the newly introduced Places148-corrupted dataset. Section 5 analyzes the results of our experiments, and finally, Section 6 presents our conclusions.

%First, Second,  Finally, 

%Finally, the contributions we present in this work are two-fold: We introduce and make publicly available the Places148-corrupted benchmark, a novel dataset for indoor scene recognition on corrupted samples. We also present baseline results on Places148-corrupted, which we expect will contribute to further research in the field of indoor scene recognition.

\section{Related Works}
%\subsection{Indoor scene recognition}
Indoor scene recognition has been shaped by various proposals that fall under two main categories: image-based classification and multimodal techniques that incorporate additional data types, such as depth maps or 3D reconstructions \citep{ 10.48550/arxiv.1911.00155, 10.48550/arxiv.2009.11154}. This discussion will, however, focus on methods that require only images as inputs, instead of other modalities, which align with the current scope of our research.

%This discussion will, however, focus on image-based methods, which align with the scope of our current research.

{
%\color{blue}The progression of image-based techniques has followed a steady pace over the years, leading to significant improvements, particularly in addressing challenges such as ambiguity and identifying distinct local spatial structures amidst highly variable scenes. An early method by \citet{khan2016discriminative} introduced a pioneering approach that targets these challenges. Their proposed model extracts dense mid-level patches from images and represents them using activations from CNNs. These patches capture a balanced level of detail and the importance of local cues with the overall scene structure. However, this approach also preserves global spatial context, which, despite its benefits, can be disadvantageous in this scenario due to the significant variability. To mitigate this, the technique incorporates supervised learning of scene patches, enhancing the preservation of local features. 

Over the years, researchers have focused their efforts on developing approaches that focus on representation learning. \citet{heikel2022indoor} proposes using object detection and Term Frequency-Inverse Document Frequency (TF-IDF), a concept borrowed from natural language processing. This method begins by training object detection models to identify items within indoor scenes and then employs TF-IDF to convert these detections into feature vectors. This vector emphasizes the most indicative objects for each scene category. A classifier is trained on these feature vectors and can achieve significant accuracy. However, this model's reliance on object detection introduces potential biases and dependencies on the object variety present in the dataset and is vulnerable to misclassification, especially in corrupted samples. Although these methodologies present noteworthy strategies for indoor scene recognition, they share a standard limitation in their efficacy when applied to corrupted samples.}

%\subsection{Perception tasks on corrupted images}
Most research focuses on removing the corruption before using the image as input. \citet{milyaev2017towards}, for example, focus on noisy images captured by surveillance cameras. Their method uses bilateral filtering, which allows for the preservation of edges while keeping image structure. Their results show that this denoising step increases detection performance over multiple methods.

Low-light images also impose a challenging scenario due to the high noise and bad contrast this corruption adds. \citet{chen2021exploring} proposes an approach to improve the accuracy of object detection models in these scenarios. The authors present a two-pronged approach. First, they enhance the low-light images using Cycle-GAN \citep{zhu2017unpaired}, and then apply object detection techniques to these enhanced images.

%Other approaches are more sophisticated and employ adversarial training in their pipelines. \cite{kim2023robust} focus on the scenario of autonomous vehicles and how harsh conditions can impact their perception systems and, therefore, impact their ability to perform autonomous planning. A problem, however, when applying adversarial training is called "catastrophic forgetting." In this case, it happens when the model learns to operate in these harsh conditions but forgets how to operate in clean conditions. The authors propose an adversarial defense module that extracts intermediate representative features to handle multi-type corruption while keeping the performance under clean conditions.

%However, these approaches add constraints related to the prior knowledge of which corruption will be present in the images to adapt the pipeline for that specific challenge. These constraints are not ideal for in-the-wild scenarios since it is difficult to estimate these parameters priorly. Other techniques have been proposed to tackle multiple corruptions at the same time.

Data augmentation is also a significant approach to improve robustness against corruptions. As suggested by \citet{wang2023robustness}, we can group these augmentation strategies into four groups: (1) Basic data augmentations, in which simple transformations are applied, such as adding color jittering or using kernel filters to apply blur and other corruptions. (2) Advanced augmentation strategies, combining different transformations to increase variability. Examples are AutoAugment \citep{cubuk2019autoaugment}, a technique that combines different transformations to increase variability through reinforcement learning and AugMax \citep{wang2021augmax}, a technique that selects augmentations based in adversarial learning. (3) Strategies based on generative models, such as GANs and VAEs. An example is the work by \citet{michaelis2019benchmarking} that uses neural style transfer to generate stylized data containing natural distortions. (4) Adversarial strategies such as adversarial noise propagation \citep{liu2021training} that injects noise to the hidden layers of the network, and diverse augmentation-based joint adversarial training \citep{addepalli2022efficient}, where two models are trained with default and augmented images. These models share the same weights, but have different statistics for batch normalization layers and are more robust to distribution shifts.

Besides image augmentation, another possibility is to mitigate corruptions through model adaptation. \citet{schneider2020improving} replaces the batch normalization statistics with those learned during training on corrupted images, leading to improvement in robustness.  \citet{chen2021robust} uses adversarial examples to adaptively select the stronger sample at each step, improving robustness against natural corruptions and domain shift.

% For gender recognition on corrupted images, \cite{greco2021gender} proposes a deep evaluation of multiple models on two datasets, reporting the performance of each model over the selected corruptions. However, they do not propose any approach to mitigate these challenging scenarios, providing insights about the performance of each model.

%Also for the description of human behavior, \cite{costa2021investigation} evaluates how challenging scenarios impact the pose of 2D pose estimation, focusing on hand pose estimation. They show that by using depthwise separable convolutions, the learned filters are more robust to corruptions while also decreasing the inference time of the model. Their proposed model can make robust predictions even when a black box occludes half of the region of the hand.

% Finally, \cite{tomy2022fusing} proposes an approach for sensor fusion, using event-based cameras and common cameras for object detection in adverse conditions. A voxel grid representation encodes event data from the event camera. At the same time, RGB images are transformed to a viewpoint using homography. A feature pyramid network architecture fuses features from both sensors at multiple scales. Their evaluation shows that this approach increases robustness in common corruptions.

Although working well on a diverse set of corruptions, these techniques share the same limitation: they all need to use corruption information during runtime. Our approach differs from these related works by not requiring any previous knowledge of the corruptions that could be present on the scene.

%Although being able to work on a diverse set of corruptions without knowing which corruption will be present a priori, these techniques also add constraints related to needing to use corruption information during runtime, being through augmentation, adaptation of augmentation, or even seeing the data in its entirety. Our approach differs from the state-of-the-art by not requiring knowledge of the corruptions a priori.

\section{Methodology}

\begin{figure*}[h!]
\centering
\includegraphics[width=\linewidth]{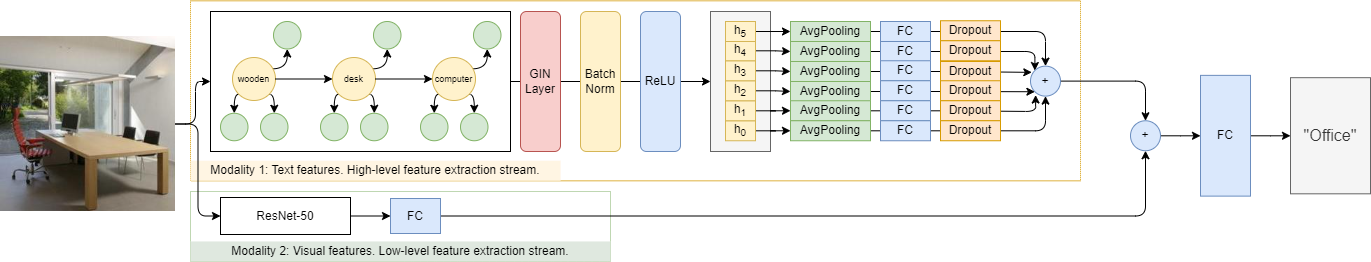}
\caption{Our proposed architecture for this study. Given an RGB image as input, we divide our execution into two branches: a \textit{high-level feature extraction stream}, in which we caption the image and create a large knowledge graph that we input to a GIN-like architecture, and a \textit{low-level feature extraction stream}, in which we use a ResNet-50 \citep{he2015deep} model to extract features. Finally, we concatenate the output of these two streams in a late fashion and apply a Fully Connected layer (FC) to do classification.}
\label{fig:arch}
\end{figure*}

Given an image \( I \), we aim to classify the indoor scene \( y \) depicted in this image. The overview of our framework is depicted in \autoref{fig:arch}. Our proposed architecture extracts features at two levels: a high-level description of the scene, denoted as \( h \), is generated using a captioning approach, and a low-level description, denoted as \( l \), is generated using Convolutional Neural Networks (CNNs). We later perform a multimodal fusion step, combining these two representations in a late fusion design.

\subsection{Extraction of high-level image descriptors}

We will describe the adaptation of high-level image descriptor extraction techniques, originally proposed by \citet{de2023high} for emotion recognition, to the domain of indoor scene recognition. First, we apply ExpansionNet-v2 \citep{hu2022expansionnet}, an image captioning model, to generate raw captions for each image in our dataset. Next, we use \textit{spacy} to remove stop words from these raw captions. Stop words are prevalent words that do not add semantic meaning to the scene, such as articles, prepositions, and pronouns.

We empirically choose to remove these words from the captions since maintaining them would elevate complexity, given their high frequency in the English language. We also remove common nouns, such as \textit{man}, \textit{woman}, \textit{girl}, and \textit{boy}, so that the model would not learn correlations between these words, as it could impact fairness. Finally, we apply lemmatization to reduce each word to its root form. We call them valid words (\(w\)), and it will be used in the following steps to generate data representations:

%\begin{enumerate}[label={(\arabic*)}]

\paragraph{Co-occurrence mining.} We proceed to generate co-occurrence matrices through co-occurrence mining. These matrices represent patterns of labels within the dataset, which will be considered through conditional probability. We count the occurrence of pairs of \((w, y)\) for each valid word \(w\) and scene label \(y\), resulting in a matrix \(M_y \in \mathbb{N}^{W \times Y}\). Therefore, \(M_{y_{ij}}\) denotes the number of times that scene \(s_j\) occurred when the valid word \(w_i\) also occurred. This is called the scene co-occurrence matrix. Following the same process, we generate a co-occurrence matrix based on the co-occurrence of valid words. Given a window of size \(s\), we slide this window to capture the co-occurrence of valid words, leading to the matrix \(M_w \in \mathbb{N}^{W \times W}\), in which \(M_{w_{ij}}\) denotes the number of times that the valid word \(w_i\) appeared together with the valid word \(w_j\). Finally, we use the valid words and the generated co-occurrence matrices to construct a global knowledge graph. Although some of the knowledge is learned prior, the definition and construction of the graph is done as needed and in real-time. This allows this technique to generate representations from unseen data.

\paragraph{Knowledge graph generation.} We start by constructing an empty graph \(G = (V, E)\), in which \(V\) is the set of nodes and \(E\) is the set of edges. In this case, \(V = E = \varnothing\). For each valid word \(w\), we start by adding a new node \(V_{w_i}\). We use GloVe \citep{pennington2014glove} to fetch the embedding of the word, using this representation as the feature \(X \in \mathbb{R}^{50}\) for this node. If the valid word is absent from GloVe, we randomly sample this embedding from a uniform distribution [-0.01, 0.01]. Next, we add a node \(V_{s_j}\) for each possible scene category in the dataset. Finally, we add edges \(e = (V_{w_i}, V_{s_j})\) that connect each valid word and each possible scene category. These edges are weighted according to the equation below:

\begin{equation}
    e_w = P(s_j | w_i) = \frac{M_{y_{w_i}}}{\sum_{k}M_{y_{w_k}}}
\end{equation}

\paragraph{Deep GCN for high-level description.} We employ a Deep Graph Convolutional Network (GCN) to generate high-level descriptions \(h\). For this task, we propose adapting the Graph Isomorphism Network (GIN) \citep{xu2018powerful}, which we chose due to its simple architecture, which could lead to reasonable inference rates in low-energy, low-consumption devices.

Inside our model, we employ a hidden representation stack \( h_s = [h_0, h_1, \ldots, h_5] \) that stores features as they are processed. Given a graph \(G\) as input, we directly store this graph's features into \(h_0\). We loop these features through a GIN convolutional block composed of a GIN layer, batch normalization, and ReLU \citep{agarap2018deep} five times, generating representations \(h_1\) to \(h_5\). We finally iterate over the hidden representations, average-pooling these features and reducing their dimensionality to \(148\). Finally, we apply a dropout of \(0.5\) to these representations and concatenate them to generate our high-level description \(h\).

\subsection{Extraction of low-level image descriptors}
We employ ResNet-50 \citep{he2015deep} as a backbone model for this stream. Its robust model architecture allows for extracting complex patterns and robust visual features. We change the last layer of this model to produce a 128-dimensional feature map from an input image. This feature map is our low-level representation \(l\) that will be used in the fusion process as described below.

\subsection{Descriptor fusing}
To enhance the accuracy and robustness of indoor scene recognition on corrupted images, we propose the fusion of high-level descriptions, which are employed in this work as captions, and low-level descriptions, employed as visual features extracted from CNNs. Although early fusion methods have generally demonstrated superior performance over late fusion approaches \citep{early_fusion1, early_fusion2}, we chose the latter, driven by the significant differences between the domains of the data. %Also, designing an early fusion approach would be challenging due to the interaction between the features we employed in earlier stages.

We concatenate the high-level description \( h \) and the low-level description \( l \) to form a combined feature vector \( z \), expressed as \( z = h \ || \ l \), in which \( || \) is the concatenation operator, thus producing a 256-dimensional output that we reduce using fully connected layers to produce an output of 148-dimensional shape. We first train the high-level stream and low-level stream independently. After this stage, we freeze their weights and allow the fusion layers to learn their representations. This approach was chosen empirically.

\section{Experiments}
% in this work we evaluate the model in this dataset extended with corruptions
In this section, we will describe the experimentation process to evaluate our model. Overall, we will discuss the construction of our extended dataset with image corruptions, as well as the definition of the metric for evaluating corruption.

\begin{figure}[t!]
\captionsetup[subfigure]{labelformat=empty}
    \centering
    \begin{subfigure}{0.20\columnwidth}
    \centering
    \includegraphics[width=\columnwidth]{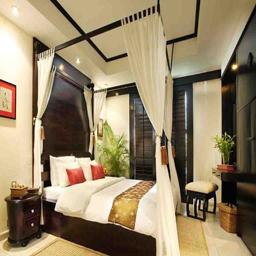}
    \caption{Bedchamber}
    \end{subfigure}
    \begin{subfigure}{0.20\columnwidth}
    \centering
    \includegraphics[width=\columnwidth]{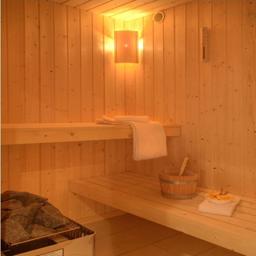}
    \caption{Sauna}
    \end{subfigure}
    \begin{subfigure}{0.20\columnwidth}
    \centering
    \includegraphics[width=\columnwidth]{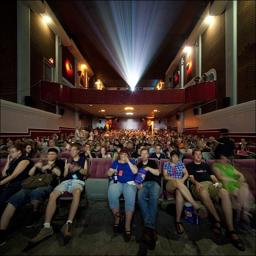}
    \caption{Movie theater}
    \end{subfigure}
    \begin{subfigure}{0.20\columnwidth}
    \centering
    \includegraphics[width=\columnwidth]{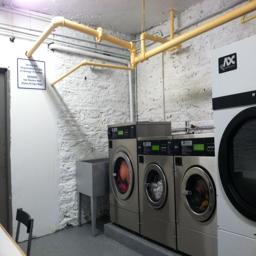}
    \caption{Utility room}
    \end{subfigure}
    
    \begin{subfigure}{0.20\columnwidth}
    \centering
    \includegraphics[width=\columnwidth]{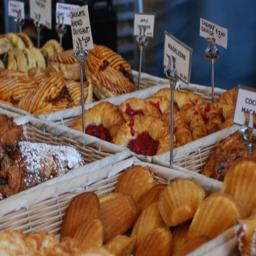}
    \caption{Bakery-shop}
    \end{subfigure}
    \begin{subfigure}{0.20\columnwidth}
    \centering
    \includegraphics[width=\columnwidth]{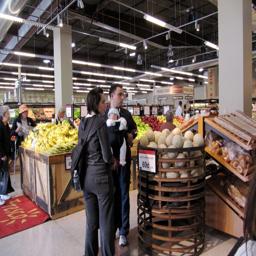}
    \caption{Supermarket}
    \end{subfigure}
    \begin{subfigure}{0.20\columnwidth}
    \centering
    \includegraphics[width=\columnwidth]{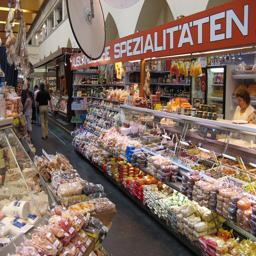}
    \caption{Market-indoor}
    \end{subfigure}
    \begin{subfigure}{0.20\columnwidth}
    \centering
    \includegraphics[width=\columnwidth]{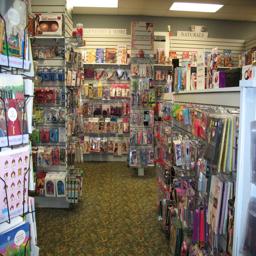}
    \caption{Gift-shop}
    \end{subfigure}
    
    \begin{subfigure}{0.20\columnwidth}
    \centering
    \includegraphics[width=\columnwidth]{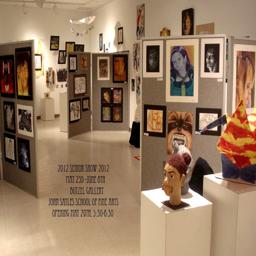}
    \caption{Art-gallery}
    \end{subfigure}
    \begin{subfigure}{0.20\columnwidth}
    \centering
    \includegraphics[width=\columnwidth]{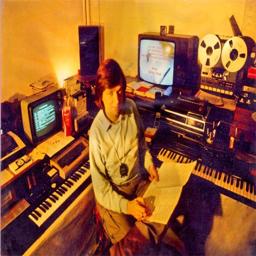}
    \caption{Music-studio}
    \end{subfigure}
    \begin{subfigure}{0.20\columnwidth}
    \centering
    \includegraphics[width=\columnwidth]{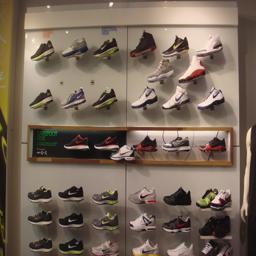}
    \caption{Shoe-shop}
    \end{subfigure}
    \begin{subfigure}{0.20\columnwidth}
    \centering
    \includegraphics[width=\columnwidth]{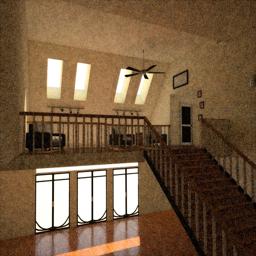}
    \caption{Mezzanine}
    \end{subfigure}
    
    \caption{Examples of selected indoor classes from Places365 dataset.}
    \label{fig:subplaces}
    \end{figure}

\subsection{Places148-corrupted: An indoor scenes dataset} We evaluate our proposed approach for indoor scene recognition on Places148, which is a subset from Places365 \citep{zhou2017places}. Places365 contains 1.8 million training images and 36,500 validation images across 365 scene classes. From these scene classes, we identified and selected 148 classes representing indoor scenes, and we utilize this subset in this study. We show samples of these classes on \autoref{fig:subplaces}.

The motivation behind this new dataset is that, although Places365 is a comprehensive and widely recognized dataset, it is focused on high-quality images without corruptions. Degradation of image quality is a frequent issue in real-world data acquisition. % Such degradation can arise from multiple factors: it may be user-induced, as in cases where images are captured out of focus or with motion blur due to unsteady handling or lack of stabilization features. Alternatively, the degradation might be environmental, stemming from inadequate lighting conditions introducing noise into the image. Additionally, technical processes, such as lossy JPEG compression, can also lead to a deterioration of image integrity. 
Therefore, we construct test sets corrupted with various types of image degradations that happen in real environments on Places148 with 15 common corruptions over five severity levels, generating a total of 75 subsets, plus one copy of the subset without any corruption. In this work, we use the well-established list of corruptions proposed by \citet{corruptions}. While certain corruptions such as \textit{Snow} and \textit{Frost} may not manifest in indoor scenarios, we have opted to include the entire spectrum of corruptions. This decision ensures a comprehensive evaluation of the model across diverse conditions, emphasizing the robustness of our proposal.

\begin{figure}[h!]
\captionsetup[subfigure]{labelformat=empty}
\centering
\begin{subfigure}{0.15\columnwidth}
\centering
\includegraphics[width=\columnwidth]{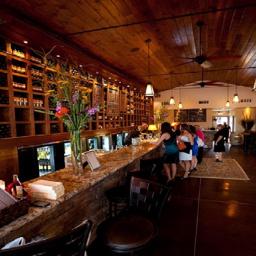}
\caption{Clean}
\end{subfigure}
\begin{subfigure}{0.15\columnwidth}
\centering
\includegraphics[width=\columnwidth]{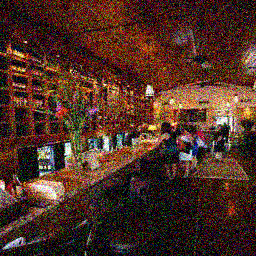}
\caption{Gauss. noise}
\end{subfigure}
\begin{subfigure}{0.15\columnwidth}
\centering
\includegraphics[width=\columnwidth]{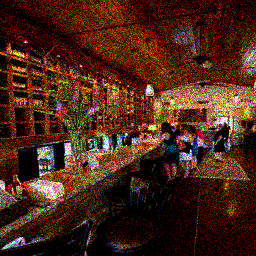}
\caption{Shot}
\end{subfigure}
\begin{subfigure}{0.15\columnwidth}
\centering
\includegraphics[width=\columnwidth]{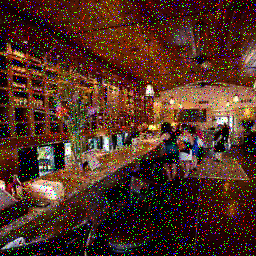}
\caption{Impulse}
\end{subfigure}
\begin{subfigure}{0.15\columnwidth}
\centering
\includegraphics[width=\columnwidth]{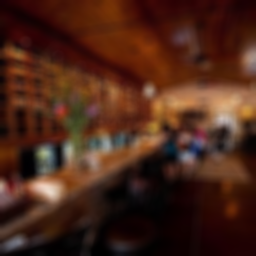}
\caption{Defocus}
\end{subfigure}

\begin{subfigure}{0.15\columnwidth}
\centering
\includegraphics[width=\columnwidth]{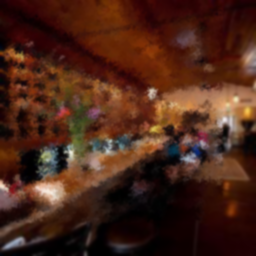}
\caption{Glass}
\end{subfigure}
\begin{subfigure}{0.15\columnwidth}
\centering
\includegraphics[width=\columnwidth]{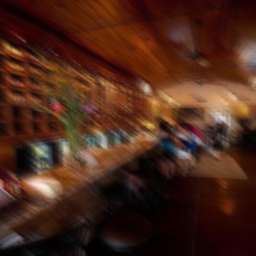}
\caption{Motion}
\end{subfigure}
\begin{subfigure}{0.15\columnwidth}
\centering
\includegraphics[width=\columnwidth]{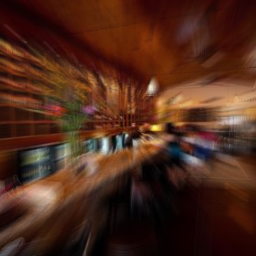}
\caption{Zoom}
\end{subfigure}
\begin{subfigure}{0.15\columnwidth}
\centering
\includegraphics[width=\columnwidth]{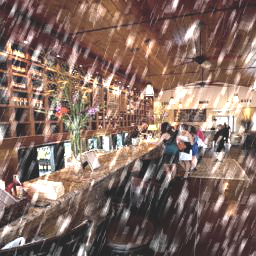}
\caption{Snow}
\end{subfigure}
\begin{subfigure}{0.15\columnwidth}
\centering
\includegraphics[width=\columnwidth]{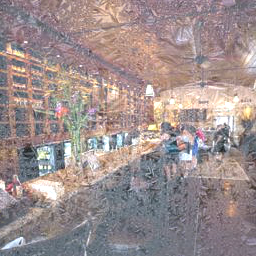}
\caption{Frost}
\end{subfigure}

\begin{subfigure}{0.15\columnwidth}
\centering
\includegraphics[width=\columnwidth]{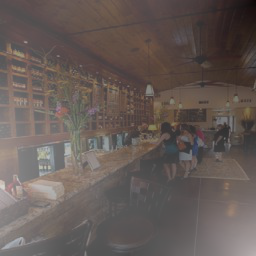}
\caption{Fog}
\end{subfigure}
\begin{subfigure}{0.15\columnwidth}
\centering
\includegraphics[width=\columnwidth]{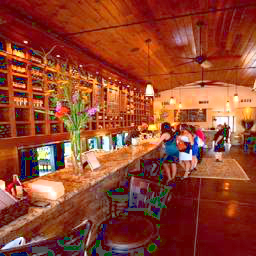}
\caption{Bright}
\end{subfigure}
\begin{subfigure}{0.15\columnwidth}
\centering
\includegraphics[width=\columnwidth]{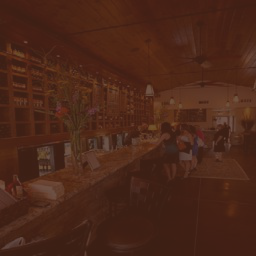}
\caption{Contrast}
\end{subfigure}
\begin{subfigure}{0.15\columnwidth}
\centering
\includegraphics[width=\columnwidth]{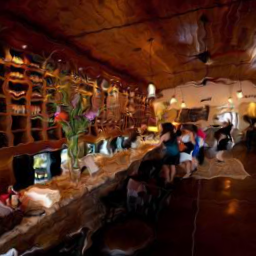}
\caption{Elastic}
\end{subfigure}
\begin{subfigure}{0.15\columnwidth}
\centering
\includegraphics[width=\columnwidth]{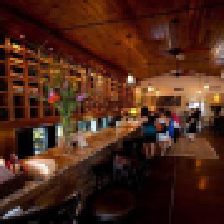}
\caption{Pixel}
\end{subfigure}
\caption{Samples of each corruption added to the dataset, with severity level \(3\). The 15 corruptions are based on the work presented in \cite{corruptions}.}
\label{fig:corruptions}
\end{figure}

We call this dataset Places148-corrupted; we show samples of each of the 15 common corruptions on a moderate severity level (\(s\) = 3) in \autoref{fig:corruptions}. A detailed description of the 148 selected categories, as well as the code for generating this dataset with the proposed corruptions are available in the repository for this project\footnote{To be included upon publication}.
\begin{comment}
\begin{table}
\caption{Based on the work by \cite{corruptions}, Places-148 is composed of 15 corruptions. We show visual examples in \autoref{fig:corruptions}.}

\begin{tabular}{ll}
\begin{tabular}[c]{@{}l@{}}1) \textbf{Gaussian noise}: Common in low-lighting conditions.\\ 2) \textbf{Shot noise} (or Poisson noise): Models how photons are acquired, \\ given their randomness depending on the illumination. \\ 3) \textbf{Impulse noise}: A color analogue of Salt \& Pepper; \\ caused by defects in the camera sensor or bit errors.\\ 4) \textbf{Defocus blur}: Happens when the image is out of blur.\\ 5) \textbf{Glass}: Appears in frosted glass, more common in outdoor environments.\\ 6) \textbf{Motion blur}: Appears when the camera is moving quickly during capture.\\ 7) \textbf{Zoom blur}: Occurs when a camera moves toward an object rapidly.\\ 8) \textbf{Snow}: Precipitation that visually obstructs content.\end{tabular} & \begin{tabular}[c]{@{}l@{}}9) \textbf{Frost}: Forms when lenses are coated with ice.\\ 10) \textbf{Fog}: Weather condition that shrouds objects.\\ It is rendered with the diamond-square algorithm.\\ 11) \textbf{Bright}: Varies with daylight intensity.\\ 12) \textbf{Contrast}: Changes depending on lighting condition.\\ 13) \textbf{Elastic}: Transformations that stretch or contract image regions.\\ 14) \textbf{Pixelation}: Occurs when upsampling low-resolution images.\\ 15) \textbf{JPEG}: A lossy image compression format that introduces \\ compression artifacts.
\end{tabular}
\end{tabular}
\label{table:corruptions}
\end{table}
\end{comment}

\textbf{Evaluating corruption.} We employ the well-known Mean Average Precision (mAP) metric, jointly with Mean Corruption Error (mCE) and Mean Relative Corruption Error (mRCE) metrics, as introduced by \cite{corruptions}, which are metrics for evaluating robustness against corruptions. 

While Corruption Error is a standardized measure that quantifies the absolute corruption error, the Relative Corruption Error assesses how much a classifier's error rate increases when processing corrupted inputs compared to clean inputs. Considering that different corruption types pose varying difficulty levels when evaluating a classifier's performance, the errors of the AlexNet model \citep{krizhevsky2012imagenet} are utilized to standardize this performance measure. 

\begin{comment}
We define these metrics by the following equations:

\begin{equation}
    \text{CE}_{c}^{f} = \left (  \sum_{s=1}^{5} E_{s,c}^{f} \right )\Big/ \left ( \sum_{s=1}^{5} E_{s,c}^{\text{AlexNet}}\right ),
    \label{eq:CE}
\end{equation}

\begin{equation}
    \text{RCE}_{c}^{f} = \left (  \sum_{s=1}^{5} E_{s,c}^{f} - E_{clean}^f\right )\Big/ \left ( \sum_{s=1}^{5} E_{s,c}^{\text{AlexNet}} - E_{clean}^{\text{AlexNet}} \right )
\label{eq:rCE}
\end{equation}

\begin{equation}
    \text{mCE}^f = \frac{1}{15}\sum_{c=1}^{15} \text{CE}_c^f .
\label{eq:mce}
\end{equation}

In this context, $E_{s,c}^{f}$ corresponds to the top-1 error of classifier $f$ for corruption $c$ at severity level $s$ ($ 1 \leqslant s \leqslant 5$).
\end{comment}

\textbf{Training procedure.}
We have trained our high-level stream using the Adam optimizer \citep{kingma2014adam} with a learning rate of \(0.0001\) and weight decay of \(0.0006\), and our low-level stream with a learning rate of \(0.001\) and a weight decay of \(0.01\). We applied early stopping when no improvement in the validation accuracy was perceived, leading to training for 15 epochs, 25 epochs, and 15 epochs, respectively, for high-level stream, low-level stream, and fusion, with a batch size of \(64\). To evaluate the quantitative results of our model, we employ the accuracy metric using top-1, top-3, and top-5 accuracies, given the high variability of classes.

\begin{table}[h]

\caption{Comparison of accuracy metrics of our proposed networks evaluated on Places148-clean. \( h \) stands for high-level descriptions only, \( l \) stands for low-level descriptions only, and \( z \) stands for the fusion of both descriptions.}\label{tab1}%

\begin{tabular}{@{}lllllllllll@{}}
\toprule
& & Accuracy & \(h\)  & \(l\)\footnotemark[1] & \(l\)\footnotemark[2] & \(z\)\footnotemark[1] & \(z\)\footnotemark[2] & & \\
\midrule
& & Top-1    & 31.44   & 47.39  & 48.76 & 54.32 & 54.72 & &  \\
& & Top-3    & 51.62   & 69.64  & 71.89 & 78.86 & 77.12 & & \\
& & Top-5    & 60.94   & 78.14  & 80.53\footnotemark[2] & 84.56 & 85.13 & & \\
\botrule
\end{tabular}
\footnotetext[*]{Using MobileNetV3.}
\footnotetext[\dagger]{Using ResNet-50.}
\label{tab:accuracy_comparison}
\end{table}

\section{Results and discussion}

\textbf{Quantitative analysis on clean dataset.} We experiment with our method on the Places148-clean, establishing a baseline that describes how these models work on uncorrupted images. We show these results on \autoref{tab:accuracy_comparison}. This experiment reveals that using only high-level descriptions \(h\) is not a good approach for the task. Given how this approach is based on captioning, different scenes can share objects and contexts among them, which could lead to confusion. A chair in a bar and a chair in a repair shop do not share any context. Relying only on low-level descriptions led to a better result; for this experiment, we evaluated two common encoder networks: ResNet-50 \citep{he2015deep} and MobileNetV3 \citep{howard2019searching}.  Combining high-level and low-level descriptions led to the best result of this evaluation, combining our captioning approach with ResNet-50.

\begin{figure}[!b]
\centering
\includegraphics[width=0.5\columnwidth]{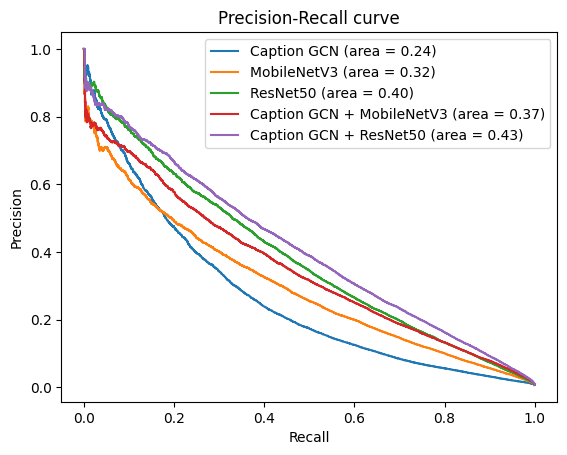}
\caption{Precision-recall curves of our proposed networks. The area under the curve (AUC) represents the mean macro-averaged precision score. Caption GCN is equivalent to \(h\), while the two CNN backbones are equivalent to \(l\), and finally the fusion being equivalent to \(z\).}
\label{figure:pr}
\end{figure}

We also evaluate the performance of the models using a precision-recall curve. We evaluate the curves of the models to ascertain their classification performance, as we show on \autoref{figure:pr}. The area under the precision-recall curve represents the model's capability of the model to generate predictions with low false positive and false negative rates. As indicated by the quantitative accuracy above, using only the low-level descriptions \(l\) generated by the Caption GCN model (GIN) demonstrates a lower area, suggesting limited effectiveness, while models that combine low-level and high-level have a higher area. Again, using ResNet-50 led to the best results.\\

\begin{figure}[t!]
    % Row with column labels
    \centering
    \makebox[0.18\columnwidth]{Input}
    \makebox[0.18\columnwidth]{\( l^\ast \)}
    \makebox[0.18\columnwidth]{\( z^\ast \)}
    \makebox[0.18\columnwidth]{\( l^\dagger \)}
    \makebox[0.18\columnwidth]{\( z^\dagger \)}
    \par % Create a new paragraph to ensure the labels and images are on separate lines

    \begin{subfigure}{\columnwidth}
        \centering
        \includegraphics[width=0.18\columnwidth]{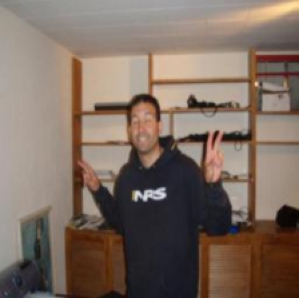}
        \includegraphics[width=0.18\columnwidth]{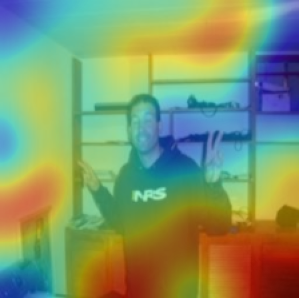}
        \includegraphics[width=0.18\columnwidth]{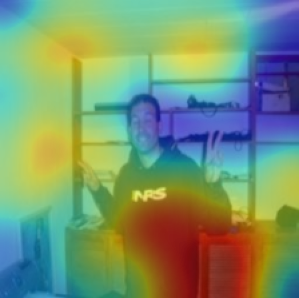}
        \includegraphics[width=0.18\columnwidth]{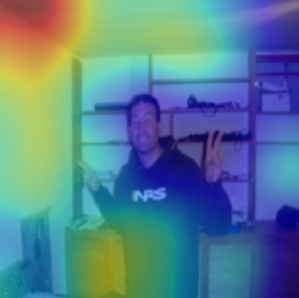}
        \includegraphics[width=0.18\columnwidth]{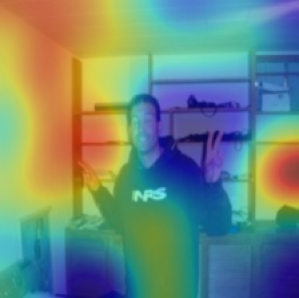}
        \caption{dorm\_room}
        \label{fig:overall:a}
    \end{subfigure}

    \begin{subfigure}{\columnwidth}
        \centering
        \includegraphics[width=0.18\columnwidth]{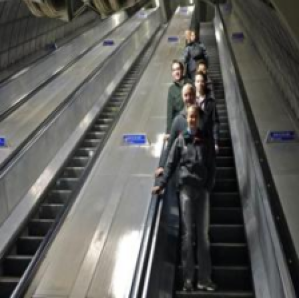}
        \includegraphics[width=0.18\columnwidth]{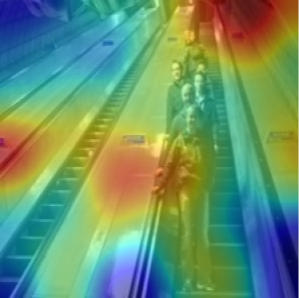}
        \includegraphics[width=0.18\columnwidth]{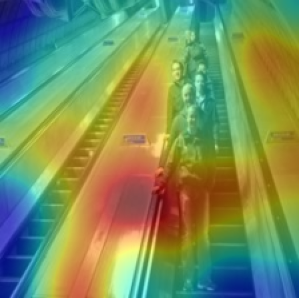}
        \includegraphics[width=0.18\columnwidth]{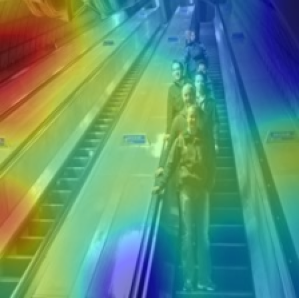}
        \includegraphics[width=0.18\columnwidth]{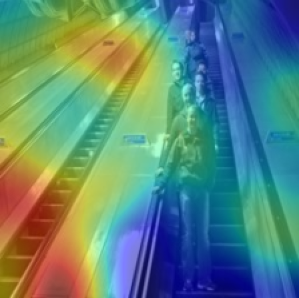}
        \caption{escalator\_indoor}
        \label{fig:overall:b}
    \end{subfigure}

    \begin{subfigure}{\columnwidth}
        \centering
        \includegraphics[width=0.18\columnwidth]{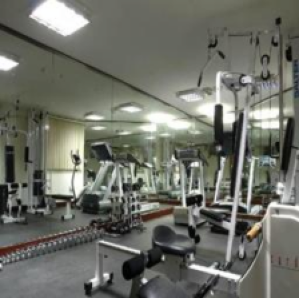}
        \includegraphics[width=0.18\columnwidth]{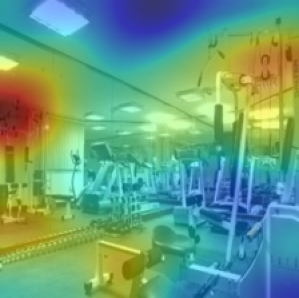}
        \includegraphics[width=0.18\columnwidth]{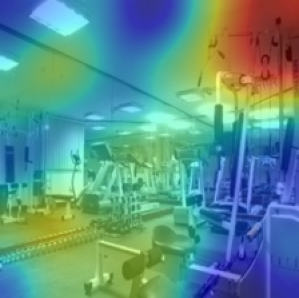}
        \includegraphics[width=0.18\columnwidth]{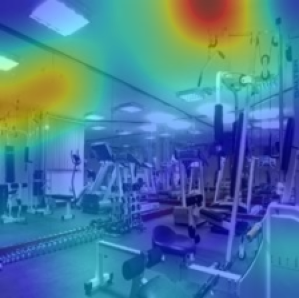}
        \includegraphics[width=0.18\columnwidth]{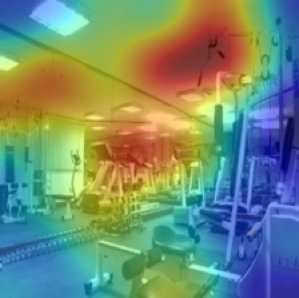}
        \caption{gymnasium\_indoor}
        \label{fig:overall:c}
    \end{subfigure}

    \begin{subfigure}{\columnwidth}
        \centering
        \includegraphics[width=0.18\columnwidth]{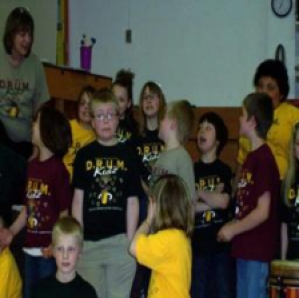}
        \includegraphics[width=0.18\columnwidth]{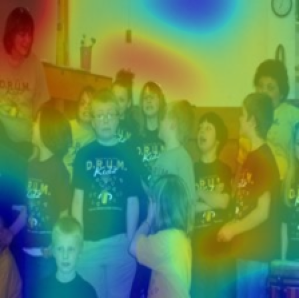}
        \includegraphics[width=0.18\columnwidth]{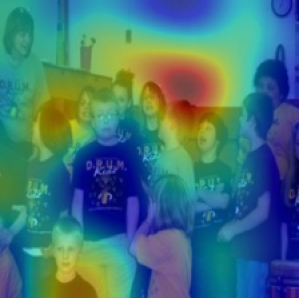}
        \includegraphics[width=0.18\columnwidth]{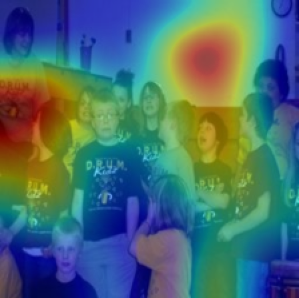}
        \includegraphics[width=0.18\columnwidth]{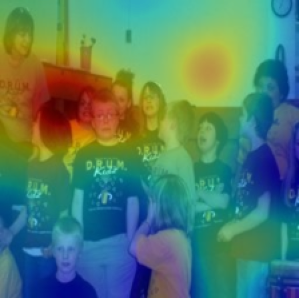}
        \caption{kindergarden\_classroom}
        \label{fig:overall:d}
    \end{subfigure}

    \begin{subfigure}{\columnwidth}
        \centering
        \includegraphics[width=0.18\columnwidth]{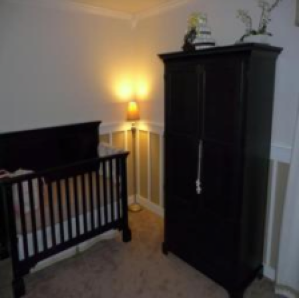}
        \includegraphics[width=0.18\columnwidth]{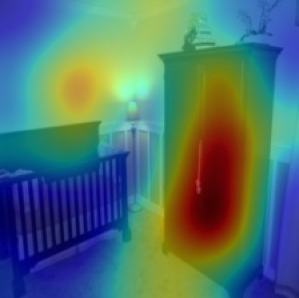}
        \includegraphics[width=0.18\columnwidth]{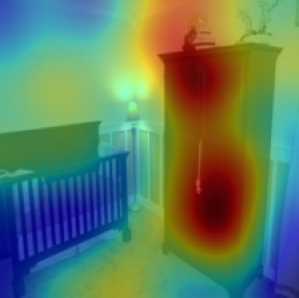}
        \includegraphics[width=0.18\columnwidth]{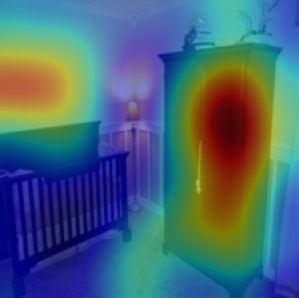}
        \includegraphics[width=0.18\columnwidth]{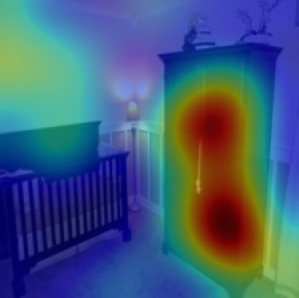}
        \caption{nursery}
        \label{fig:overall:e}
    \end{subfigure}

    \caption{Comparison of saliency maps generated by each proposed network. The first column is the input image, while each subsequent column shows the activation maps regarding the class that is captioned under each row. Regions in red and yellow are regions of higher interest for the network. \( h \) stands for high-level descriptions only, \( l \) stands for low-level descriptions only, and \( z \) stands for the fusion of both descriptions. * using MobileNetV3 and $\dagger$ using ResNet-50.
    }
    \label{fig:overall}
\end{figure}

\textbf{Qualitative analysis on clean dataset.} We also evaluate our model using Grad-CAM \citep{selvaraju2022grad}, in which we generate saliency maps that indicate regions of the images that are more important during classification. These can be interpreted as highlighted areas for that prediction. As we can see from \autoref{fig:overall}, although each backbone CNN can highlight important regions of the scene, the activations are increased when combined with our Caption GCN approach.

%We also evaluate our model with a qualitative analysis focused on saliency maps. We show on \autoref{fig:overall} five randomly selected example images from the dataset and their corresponding saliency maps for our proposed architectures. Saliency maps indicate regions of the images that are more important for the model during the classification step; they can be seen as highlighted areas for that prediction. We generate the maps using Grad-CAM \citep{selvaraju2022grad}, which works by backpropagating the gradient of the output class with respect to the input image, capturing the influence of each original region for the prediction. To improve visualization, we normalize and smooth the saliency maps with a Gaussian filter. As we can see from this evaluation, although each backbone CNN can highlight important regions of the scene, the activations are increased when combined with our Caption GCN approach.

The generated saliency maps points out to the increased density around descriptive regions of the image when using multimodal fusion models (\(z\)) in respect to using only low-level features (\(l\)). In \autoref{fig:overall:a}, which is a sample from the \textit{dorm\_room} class, we can see that \(l^\ast\) has a sparse map which highlights most of the bottom part, as well as an undescriptive region of the wall. Moving to \(z^\ast\), the saliency maps gets more dense around the furniture in the background. Moving to the models using ResNet-50 as a backbone, \(l^\dagger\) shows a focus around the ceiling, with only a shallow mapping to the furniture, as \(z^\dagger\) improves this mapping to also include regions of the furniture. The same behavior can be noticed in \autoref{fig:overall:b} and on \autoref{fig:overall:c}. For models based on MobileNetV3, the highlighted regions are somewhat descriptive, but still sparse, while when moving to the fusion model \(z^\ast\) these regions are more dense around details that we might also consider important. However, the regions highlighted by \(z^\dagger\) are also improvements, pointing to other descriptive regions on the image. 

For \autoref{fig:overall:d}, the context of the image is not at all descriptive since the children depicted on the image are occupying a significant space. In this case, one may consider that \(l^\ast\) generated the more descriptive mapping, since looking at the children might be the best description for \textit{kindergarden\_classroom}. Finally, in \autoref{fig:overall:e}, we see the same behavior as before, with fusion models (\(z\)) generating more densely mappings around furniture.\\

\begin{figure}[H]
\centering
\includegraphics[width=0.60\columnwidth]{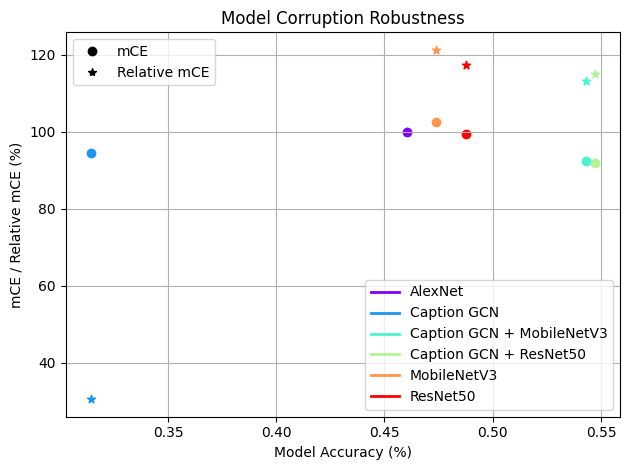}
\caption{mCE and mRCE results of each model.}
\label{fig:corruption_robustness}
\end{figure}

\textbf{Quantitative analysis on corrupted data.} We have computed RCE and mRCE metrics to evaluate the performance of each scene recognition models on images with common visual corruptions. Following \cite{corruptions}, we use an AlexNet model \citep{krizhevsky2012imagenet} trained on the clean dataset to have a baseline for that score, which is then used to standardize our scores. This approach allows for a better assessment of the corruption robustness of our proposed models since each corruption poses a different level of difficulty. We equate AlexNet scores to 100, meaning that models below this value display an improvement over the AlexNet model. This approach also allows for direct comparison between our methods. In \autoref{fig:corruption_robustness}, we show mCE and mRCE metrics with respect to each model's top-1 accuracy.

\clearpage
We also perform a comprehensive and detailed quantitative analysis on the performance variations of the different proposed models across a range of severity levels pertaining to the specific corruptions we have selected in this study. \autoref{longtable} shows the results of this evaluation. Each column represents the top-1 accuracy for a particular subset, culminating in the final column, which averages these accuracies. %This ``average" column serves a dual purpose.

%A key feature of this table is its see of visual cues: whenever a model's accuracy in a given corruption exceeds the average, the corresponding cell is highlighted. This highlighting not only draws attention but also signifies that the model's performance in that scenario is above the average, providing an intuitive and immediate understanding of which models excel under specific conditions. Overall, this comprehensive table offers a multifaceted view of model performances, allowing for nuanced comparisons and in-depth understanding of model robustness across various corruption types and severity levels.

%We also perform a comprehensive quantitative study on the performance of each model among various severity levels on each of the corruptions we proposed to investigate in this work. We show this result in \autoref{longtable}, and it is structured as follows: each column is the top-1 accuracy on this subset, and the last column is an average of these results. The "average" column can compare how multiple models perform at the same severity level or the decay in accuracy among severity levels. Each row contains the performance of each model in these subsets, and the last row of each severity grouping is the average accuracy of each corruption for that specific severity. If the accuracy of that model in that corruption is higher than the average, the cell is highlighted, indicating a higher-than-average accuracy.

\paragraph{Per-corruption analysis} Through the results exposed in \autoref{longtable}, we can see that \textit{Brightness} is the less harmful perturbation on lower severity levels, while \textit{JPEG} leads to best results on higher severity. On the other hand, \textit{Impulse} is the most harmful corruption among all models, leading to the lower average accuracy on all severities. Using only high-level features (\(h\)) leads to the best accuracy in this scenario, and using only low-level features (\(l\)) are specially prejudicial.

A curious behavior is that, although \textit{Gaussian noise} and \textit{Shot} are noises somewhat similar to \textit{Impulse}, their behavior strongly differs from this last corruption, showing similar scores among all models that fluctuate close to the average. This can be explained by how these noises are generated. Both are fundamentally random and continuous but act in a uniform manner. \textit{Gaussian noise} adds a consistent level of randomness to the pixel values across the entire image, resulting in a kind of ``static" that does not preferentially target a specific region, feature, or structure. \textit{Shot} noise is also evenly distributed, leading to a similar uniform degradation of the image quality. Both types of noise impact the image globally, but in a predictable way, without creating abrupt or localized distortions. This similarity in their behavior points out that, while they can degrade the quality of an image, they do so in a way that is often more manageable and less destructive than \textit{Impulse}.

This latter noise is characterized by its sporadic and localized nature, manifesting as sudden and sharp random fluctuations in pixel values. This irregular pattern creates a stark contrast to the underlying image content, leading to a more pronounced effect. These abrupt changes in pixel values are not representative of the true scene and can be especially challenging for image processing tasks, as they may mask key features or introduce misleading information. 

Another curious behavior is related to \textit{Fog} and its accuracies on \(l\). Consistently, \(l^\dagger\), which is the model based on ResNet-50, yields significantly improved accuracies over \(l^\ast\), the model based on MobileNetV3. This distinction can be primarily attributed to their design. ResNet-50 has a depth and complexity that could lead to an advantage in this corruption that generally reduces contrast and blurs details in the images since it can rely on more subtle feature extraction processes to make accurate predictions. MobileNetV3, on the other hand, being optimized for efficiency and speed, has a limited ability to extract sufficient relevant features in these obscured images. As expected, ResNet-50 also yields higher accuracies on average for all corruptions, except for \textit{Elastic} on severities 4 and 5.

\begin{sidewaystable*}
    \centering
    \caption{Evaluation of performance of the proposed models across a spectrum of corruption types and severity levels. \( h \) stands for high-level descriptions only, \( l \) stands for low-level descriptions only, and \( z \) stands for the fusion of both descriptions.}
    \renewcommand{\arraystretch}{1.3}
\resizebox{.95\columnwidth}{!}{%
\begin{tabular}{clp{1cm}p{1cm}p{1cm}p{1cm}p{1cm}p{1cm}p{1cm}p{1cm}p{1cm}p{1cm}p{1cm}p{1cm}p{1cm}p{1cm}p{1cm}p{1cm}}
    \hline
    \multirow[t]{2}{*}{Severity} & \multirow[t]{2}{*}{Model} & \multicolumn{15}{l}{Corruption (top-1 acc. in \%)}                                                          & Avg\\    
    &                        & Gauss. & Shot. & Impul. & Defoc. & Glass & Motion & Zoom & Snow & Frost & Fog  & Bright & Contr. & Elast. & Pixel & JPEG &      \\ 
    \hline
    \multirow{5}{*}{1}        & \(h\)                        & \colorbox{gray!25}{29.7} & \colorbox{gray!25}{29.7} & \colorbox{gray!25}{29.7} & \colorbox{gray!25}{28.0} & 27.8 & 29.6 & 24.3 & \colorbox{gray!25}{28.3} & 29.0 & 29.7 & 31.5 & 30.2 & 29.4 & 29.7 & 30.5 & 29.1\\
                              & \(l^\ast\)                       & 20.8 & 21.1 & 16.3 & 18.5 & 28.0 & 30.8 & 21.7 & 22.5 & 27.3 & 23.0 & 45.7 & 24.3 & 39.0 & 39.3 & 44.4 & 28.2\\
                              & \(l^\dagger\)                      & 24.6 & 25.7 & 18.6 & 22.0 & 29.2 & 33.3 & 24.8 & 23.0 & 30.6 & \colorbox{gray!25}{31.7} & \colorbox{gray!25}{47.5} & \colorbox{gray!25}{32.8} & 40.2 & \colorbox{gray!25}{44.0} & \colorbox{gray!25}{45.5} & 31.6\\
                              & \(z^\ast\)                       & \colorbox{gray!25}{32.0} & \colorbox{gray!25}{31.4} & \colorbox{gray!25}{27.7} & \colorbox{gray!25}{29.7} & \colorbox{gray!25}{37.4} & \colorbox{gray!25}{41.2} & \colorbox{gray!25}{31.3} & \colorbox{gray!25}{32.9} & \colorbox{gray!25}{38.8} & \colorbox{gray!25}{35.3} & \colorbox{gray!25}{53.5} & \colorbox{gray!25}{36.4} & \colorbox{gray!25}{47.4} & \colorbox{gray!25}{50.1} & \colorbox{gray!25}{51.6} & \colorbox{gray!25}{38.4}\\
                              & \(z^\dagger\)                      & \colorbox{gray!25}{34.7} & \colorbox{gray!25}{34.9} & \colorbox{gray!25}{27.8} & \colorbox{gray!25}{29.4} & \colorbox{gray!25}{37.5} & \colorbox{gray!25}{41.5} & \colorbox{gray!25}{32.2} & \colorbox{gray!25}{31.9} & \colorbox{gray!25}{38.8} & \colorbox{gray!25}{39.0} & \colorbox{gray!25}{53.6} & \colorbox{gray!25}{40.5} & \colorbox{gray!25}{47.8} & \colorbox{gray!25}{49.5} & \colorbox{gray!25}{52.1} & \colorbox{gray!25}{39.4}\\
     \rowcolor{gray!7}& Avg.                         & 28.3 & 28.5 & 24.0 & 25.5 & 31.9 & 35.3  & 26.9  & 27.7  & 32.9  & 31.7  & 46.4  & 32.8 & 40.7  & 42.5  & 44.8  & 33.3\\
    \hline
    \multirow{5}{*}{2}        & \(h\)                        & \colorbox{gray!25}{27.6} & \colorbox{gray!25}{28.0} & \colorbox{gray!25}{28.2} & \colorbox{gray!25}{26.0} & \colorbox{gray!25}{25.2} & \colorbox{gray!25}{27.2} & 20.1 & \colorbox{gray!25}{22.0} & \colorbox{gray!25}{25.2} & \colorbox{gray!25}{29.1} & 31.1 & \colorbox{gray!25}{29.6} & 19.2 & 29.4 & 30.1 & \colorbox{gray!25}{26.5}\\
                              & \(l^\ast\)                       & 9.8 & 9.6 & 5.9 & 12.6 & 15.5 & 19.6 & 16.2 & 9.1 & 12.5 & 17.3 & 41.3 & 15.5 & 24.0 & 39.3 & 42.9 & 19.4\\
                              & \(l^\dagger\)                      & 14.1 & 14.7 & 9.4 & 15.0 & 17.6 & 22.8 & 19.1 & 10.0 & 15.3 & 26.2 & 43.0 & 24.9 & 24.3 & \colorbox{gray!25}{44.1} & \colorbox{gray!25}{43.9} & 22.9\\
                              & \(z^\ast\)                       & \colorbox{gray!25}{19.4} & 18.1 & 14.7 & \colorbox{gray!25}{22.8} & \colorbox{gray!25}{24.5} & \colorbox{gray!25}{31.5} & \colorbox{gray!25}{24.6} & \colorbox{gray!25}{16.8} & \colorbox{gray!25}{22.9} & \colorbox{gray!25}{30.1} & \colorbox{gray!25}{50.4} & \colorbox{gray!25}{28.1} & \colorbox{gray!25}{31.3} & \colorbox{gray!25}{49.3} & \colorbox{gray!25}{50.7} & \colorbox{gray!25}{29.0}\\
                              & \(z^\dagger\)                      & \colorbox{gray!25}{22.6} & \colorbox{gray!25}{22.3} & \colorbox{gray!25}{16.9} & \colorbox{gray!25}{21.4} & \colorbox{gray!25}{25.8} & \colorbox{gray!25}{30.9} & \colorbox{gray!25}{25.3} & \colorbox{gray!25}{17.0} & \colorbox{gray!25}{23.3} & \colorbox{gray!25}{34.3} & \colorbox{gray!25}{49.9} & \colorbox{gray!25}{32.7} & \colorbox{gray!25}{31.3} & \colorbox{gray!25}{49.7} & \colorbox{gray!25}{50.6} & \colorbox{gray!25}{30.2}\\
     \rowcolor{gray!7}& Avg.                         & 18.7 & 18.5 & 15.0 & 19.5 & 21.7 & 26.4  & 21.0  & 14.9  & 19.8  & 27.4  & 43.1  & 26.1 & 26.0  & 42.3 & 43.6  & 25.4\\
    \hline
    \multirow{5}{*}{3}        & \(h\)                        & \colorbox{gray!25}{25.1} & \colorbox{gray!25}{25.5} & \colorbox{gray!25}{26.5} & \colorbox{gray!25}{20.5} & \colorbox{gray!25}{18.5} & \colorbox{gray!25}{22.7} & 16.9 & \colorbox{gray!25}{24.8} & \colorbox{gray!25}{22.2} & \colorbox{gray!25}{28.4} & 30.3 & \colorbox{gray!25}{28.3} & 24.7 & 27.7 & 29.2 & \colorbox{gray!25}{24.7}\\
                              & \(l^\ast\)                       & 3.2 & 4.2 & 2.9 & 8.1 & 7.2 & 12.1 & 13.4 & 10.8 & 6.9 & 12.8 & 34.8 & 7.0 & \colorbox{gray!25}{38.3} & 21.2 & 40.9 & 14.9\\
                              & \(l^\dagger\)                      & 5.7 & 7.0 & 5.3 & 8.9 & 8.4 & 14.6 & 15.7 & 12.8 & 9.0 & 21.6 & 36.8 & 13.8 & \colorbox{gray!25}{36.7} & 27.8 & \colorbox{gray!25}{42.8} & 17.8\\
                              & \(z^\ast\)                       & 9.3 & 10.3 & 9.1 & \colorbox{gray!25}{16.6} & \colorbox{gray!25}{12.1} & \colorbox{gray!25}{22.1} & \colorbox{gray!25}{20.8} & \colorbox{gray!25}{19.4} & \colorbox{gray!25}{15.6} & \colorbox{gray!25}{25.4} & \colorbox{gray!25}{45.2} & \colorbox{gray!25}{18.1} & \colorbox{gray!25}{44.5} & \colorbox{gray!25}{34.9} & \colorbox{gray!25}{49.5} & \colorbox{gray!25}{23.5}\\
                              & \(z^\dagger\)                      & \colorbox{gray!25}{11.8} & \colorbox{gray!25}{13.1} & \colorbox{gray!25}{11.5} & \colorbox{gray!25}{14.0} & \colorbox{gray!25}{14.2} & \colorbox{gray!25}{21.3} & \colorbox{gray!25}{21.0} & \colorbox{gray!25}{20.9} & \colorbox{gray!25}{15.6} & \colorbox{gray!25}{29.3} & \colorbox{gray!25}{44.5} & \colorbox{gray!25}{21.0} & \colorbox{gray!25}{43.6} & \colorbox{gray!25}{33.7} & \colorbox{gray!25}{49.2} & \colorbox{gray!25}{24.3}\\
     \rowcolor{gray!7}& Avg.                         & 11.0 & 12.0 & 11.0 & 13.6 & 12.1 & 18.6  & 17.6  & 17.7  & 13.8  & 23.5  & 38.3  & 17.6 & 37.5  & 29.1  & 42.3  & 21.0\\
    \hline
    \multirow{5}{*}{4}        & \(h\)                        & \colorbox{gray!25}{21.6} & \colorbox{gray!25}{20.7} & \colorbox{gray!25}{22.3} & \colorbox{gray!25}{15.5} & \colorbox{gray!25}{16.6} & \colorbox{gray!25}{16.8} & 14.3 & \colorbox{gray!25}{21.7} & \colorbox{gray!25}{22.1} & \colorbox{gray!25}{27.2} & 29.3 & \colorbox{gray!25}{23.2} & 19.8 & \colorbox{gray!25}{24.7} & 28.3 & \colorbox{gray!25}{21.6}\\
                              & \(l^\ast\)                       & 1.2 & 1.5 & 1.0 & 5.9 & 5.7 & 8.0 & 11.3 & 6.9 & 5.9 & 12.6 & 26.9 & 2.2 & \colorbox{gray!25}{30.0} & 10.8 & 35.0 & 10.9\\
                              & \(l^\dagger\)                      & 2.0 & 2.6 & 1.8 & 5.8 & 6.9 & 9.8 & 13.2 & 9.5 & 8.4 & 20.9 & 29.1 & 4.0 & 27.5 & 13.8 & 36.3 & 12.7\\
                              & \(z^\ast\)                       & 4.9 & 5.5 & 4.5 & \colorbox{gray!25}{11.8} & 9.8 & \colorbox{gray!25}{15.2} & \colorbox{gray!25}{17.2} & \colorbox{gray!25}{14.4} & \colorbox{gray!25}{14.6} & \colorbox{gray!25}{24.5} & \colorbox{gray!25}{38.0} & 8.2 & \colorbox{gray!25}{35.6} & \colorbox{gray!25}{20.3} & \colorbox{gray!25}{45.0} & \colorbox{gray!25}{17.9}\\
                              & \(z^\dagger\)                      & 5.5 & 5.9 & 5.1 & 9.5 & \colorbox{gray!25}{11.6} & \colorbox{gray!25}{13.9} & \colorbox{gray!25}{17.3} & \colorbox{gray!25}{16.3} & \colorbox{gray!25}{14.7} & \colorbox{gray!25}{28.5} & \colorbox{gray!25}{37.3} & 8.3 & \colorbox{gray!25}{34.8} & \colorbox{gray!25}{18.7} & \colorbox{gray!25}{43.7} & \colorbox{gray!25}{18.1}\\
     \rowcolor{gray!7}& Avg.                         & 7.04 & 7.24 & 6.94 & 9.7 & 10.1 & 12.7  & 14.6  & 13.7  & 13.1  & 22.7  & 32.1  & 9.1 & 29.5  & 17.6  & 37.6  & 16.2\\
    \hline
    \multirow{5}{*}{5}        & \(h\)                        & \colorbox{gray!25}{16.1} & \colorbox{gray!25}{16.7} & \colorbox{gray!25}{17.8} & \colorbox{gray!25}{11.0} & \colorbox{gray!25}{13.3} & \colorbox{gray!25}{12.7} & 11.4 & \colorbox{gray!25}{18.7} & \colorbox{gray!25}{19.5} & \colorbox{gray!25}{22.1} & \colorbox{gray!25}{27.8} & \colorbox{gray!25}{11.9} & 9.2 & \colorbox{gray!25}{22.3} & 26.3 & \colorbox{gray!25}{17.1}\\
                              & \(l^\ast\)                       & 0.7 & 0.9 & 0.7 & 4.6 & 4.6 & 6.3 & 10.1 & 5.6 & 4.4 & 8.5 & 18.9 & 1.2 & \colorbox{gray!25}{16.7} & 8.7 & 26.7 & 7.9\\
                              & \(l^\dagger\)                      & 1.2 & 1.7 & 1.1 & 4.6 & 5.9 & 7.7 & 11.7 & 7.6 & 5.7 & 14.9 & 21.1 & 1.6 & 14.7 & 10.1 & 27.9 & 9.1\\
                              & \(z^\ast\)                       & 2.3 & 3.5 & 2.5 & \colorbox{gray!25}{8.3} & \colorbox{gray!25}{8.3} & \colorbox{gray!25}{11.7} & \colorbox{gray!25}{14.9} & \colorbox{gray!25}{11.3} & \colorbox{gray!25}{10.9} & \colorbox{gray!25}{17.8} & \colorbox{gray!25}{29.6} & 2.8 & \colorbox{gray!25}{18.9} & \colorbox{gray!25}{15.9} & \colorbox{gray!25}{37.5} & \colorbox{gray!25}{13.1}\\
                              & \(z^\dagger\)                      & 2.3 & 3.4 & 2.4 & 6.4 & \colorbox{gray!25}{9.9} & \colorbox{gray!25}{10.6} & \colorbox{gray!25}{14.8} & \colorbox{gray!25}{12.7} & \colorbox{gray!25}{10.9} & \colorbox{gray!25}{20.9} & \colorbox{gray!25}{28.8} & 3.0 & \colorbox{gray!25}{18.7} & \colorbox{gray!25}{14.6} & \colorbox{gray!25}{35.2} & \colorbox{gray!25}{12.9}\\
     \rowcolor{gray!7}& Avg.                         & 4.5 & 5.2 & 4.9 & 6.9 & 8.3 & 9.8  & 12.5  & 11.1  & 10.3  & 16.8  & 25.2  & 4.1 & 15.6  & 14.3  & 30.7  & 12.0\\
    \hline
    \end{tabular}}
    \label{longtable}
    \end{sidewaystable*}

\paragraph{Per-severity analysis} Evaluating how different severity levels impact the performance of the models, we can see that when \(s = 1\), fusion models (\(z\)) steadily outperforms both high-level (\(h\)) and low-level (\(l\)) approaches, following the same behavior as \autoref{tab:accuracy_comparison}. In this case, \( z^\dagger \), which is the fusion model using ResNet-50 as a low-level feature extractor, performs 6\% better than the average and 1\% better than the second-rating model, \( z^\ast \), which uses MobileNet-v3 as a low-level feature extractor. 

Increasing the severity level to \(s = 2\), we see that fusion models also perform best, but \(h\) starts to increase its accuracy, and although it is not the best-performing model overall, in each scenario that its accuracy is higher than the average, this model also outperforms the fusion models. This is a consistent behavior indicating that if more severe image corruptions are expected, this approach could lead to a more robust description of the scene. Overall, the average accuracy of all models on all corruptions drops by 8\%, indicating a 23.72\% decrease in accuracy.

\begin{table}[b!]
\caption{Comparison of corruption severities on captioning generation for our high-level feature extraction \(h\).}
\centering
\begin{tabular}{llll} 
\toprule
Figure            & Corruption & Sev. & Captions  \\ 
\midrule
\multirow{5}{*}{\includegraphics[width=0.11\textwidth]{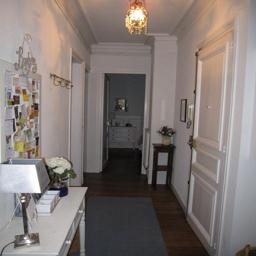}} & \multirow[t]{5}{*}{Frost}   & 1    &   A large bathroom with a sink and a door.        \\
                &     & 2    & A bathroom with a sink and a door.          \\
                &     & 3    & A bathroom with a sink and a mirror.          \\
                &     & 4    & A bathroom with a sink and a table.          \\
                &     & 5    & A bathroom with purple lights and a sink.          \\
\midrule
\multirow{5}{*}{\includegraphics[width=0.11\textwidth]{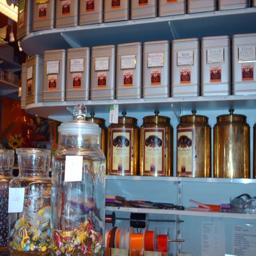}} & \multirow[t]{5}{*}{Fog}   & 1    &   A display of jars in a store.        \\
                &     & 2    & A display of jars in a store.          \\
                &     & 3    & A display of jars in a store.          \\
                &     & 4    & A store with a bunch of shelves on display.          \\
                &     & 5    & A store with a bunch of shelves on it.          \\
\midrule
\multirow{5}{*}{\includegraphics[width=0.11\textwidth]{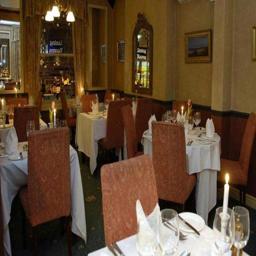}} & \multirow[t]{5}{*}{Elastic}   & 1    &   A restaurant with tables and chairs and a table.        \\
                &     & 2    & A restaurant with tables and chairs and a table.          \\
                &     & 3    & A group of people sitting at tables in a restaurant.          \\
                &     & 4    & A group of people sitting at tables in a restaurant.          \\
                &     & 5    & A group of people sitting in a room.          \\
\bottomrule
\end{tabular}
\label{table:comparison_corruptions_gcn}
\end{table}

In \(s = 3\), that is the mean term on this study, \(z^\ast\), that is our fusion model based on MobileNetV3 starts to show a significative performance drop in relation to the other models. From this point, \(h\) starts to maintain higher accuracy than the other models. Finaly, on \(s > 3\), we can see a significant dominance of \(h\) over other models. his behavior can be explained by how \(h\) extracts features from the image. Since captioning is much more dependent on context, adding corruptions may prejudice the caption generation. However, other aspects from context are still recovered, while this information is completely depleted on models that use low-level features from images. This study points to us that the usage of \(h)\) is beneficial in every case; in low-severity cases, using it by fusing with low-level models increase performance, and in high-severity cases, using it by its own points to a sufficient performance. This study also points to the low performance of models based only on low-level features. As opposed to \autoref{tab:accuracy_comparison}, these models do not present a consistent accuracy on corruptions since, as discussed above, much vital information is discarded. In \autoref{table:comparison_corruptions_gcn}, we show examples of how the caption is impacted over different severity levels. Even with high severity levels, the captions tend to maintain some descriptions of the scene.

The maximum drop in accuracy occurs from \(s = 1\) to \(s = 2\) for our fusion model using MobileNetV3 (\(z^\ast\)), dropping 9.4\%, which is equivalent to a decrease of 24.5\%. Consistently, fusion models suffers the most when increasing severity: 5.9\%, 6.2\% and 5.2\% for \(s = 2\) to \(s = 3\), \(s = 3\) to \(s = 4\) and \(s = 4\) to \(s = 5\), respectively, were the highest accuracy losses overall, which were suffered by our fusion model using ResNet-50 (\(z^\dagger\)). This evaluation points out that, even though \(h\) is unable to perform well on low severity levels, it is overall the model that is able to sustain a significant accuracy on higher severity levels, losing only 12\% accuracy, which is 14.5\% lower than the worst-performing model, \(z^\dagger\). 

\begin{comment}

We show on \autoref{table:average} a sum of the average results on all severities, on all corruptions, together with its standard deviation. This shows that, although \(z^\dagger\) is still the top performing model even when taking into account the corruptions, this value floats significantly and can be severely impacted by some corruptions.

\begin{table}[h!]
\centering
\caption{Average accuracy and standard deviation of model performance across all severity levels for the evaluated corruptions. \( h \) stands for high-level descriptions only, \( l \) stands for low-level descriptions only, and \( z \) stands for the fusion of both descriptions.}
\begin{tabular}{ccccc} 
\toprule
\(h\)            & \(l\)\footnotemark[1]         & \(l\)\footnotemark[2]     & \(z\)\footnotemark[1]         & \(z\)\footnotemark[2]     \\
\midrule
 23.82 $\pm$ 5.66 &  16.28 $\pm$ 12.31 & 18.85 $\pm$ 12.62 & 24.40 $\pm$ 13.92 & 25.00 $\pm$ 13.86  \\
\bottomrule
\end{tabular}
\footnotetext[*]{Using MobileNetV3.}
\footnotetext[\dagger]{Using ResNet-50.}
\label{table:average}
\end{table}

\end{comment}

\section{Conclusion}
%In this study, we explore the complexity of image-based indoor scene recognition under corruptions, uncovering valuable insights through quantitative and qualitative analysis. We propose three means of characterizing an image: (1) using only a high-level feature extractor, \(h\), that is based only on image captioning and graph generation; (2) using only low-level feature extractors, namely ResNet-50 and MobileNetV3, which we refer to as \(l\); and (3), performing a multimodal fusion approach to combine both low-level and high-level features, which we refer to as \(z\). We also propose a new baseline for evaluation, a subset of indoor scenes from the well-known Places365 dataset, that we refer to as Places148. Aiming to investigate how corruptions can impact this task, we also generate Places148-corrupted, which are 15 copies of Places148 with different levels of added corruptions.

Indoor scene recognition is a challenging task that is increasingly attracting attention from the computer vision community due to its application in social robotics. In this work, we propose a novel benchmark for corrupted indoor scene recognition, together with baseline results. Our proposed fusion approach combines global descriptors as low-level features and extracted captions as high-level features.

The obtained results by our models achieve significant top-1, top-3, and top-5 metrics on the clean Places148 dataset. This indicates that the proposed hybrid approach extracts a good representation of the environment. When extending this evaluation on corrupted data, we also include well-known metrics that measure corruption levels, highlighting the real-world applicability of our proposed methods. The obtained results when corruption is present also demonstrate the robustness of our proposed model in scenarios where image integrity is compromised.

We hope that the contributions of this work will pave the way for further work on corrupted scene recognition, given its potential for real work applications. Our future work will explore the effect of corruption on other data modalities, such as depth, extracted from images and videos depicting indoor scenes.
\newpage

\section*{Data Availability Statement}
The dataset generated during this research will be made publicly available at \url{https://places-corrupted.github.io} upon the publication of this manuscript.

\bibliography{sn-bibliography}% common bib file

\end{document}